# Agentic AI Optimisation (AAIO): what it is, how it works, why it matters, and how to deal with it

Luciano Floridi[1,2], Carlotta Buttaboni[1], Nicolas Gertler[1], Emmie Hine[1,2,4], Jessica Morley[1], Claudio Novelli[1], Tyler Schroder [1,3]

[1] Digital Ethics Center, Yale University, 85 Trumbull St., New Haven, CT 06511, USA

[2] Department of Legal Studies, University of Bologna, Via Zamboni 27/29, 40126 Bologna, Italy

[3] Department of Computer Science, 51 Prospect St, New Haven, CT 06511, USA

[4] Centre for IT & IP Law, Sint-Michielsstraat 6, 3000 Leuven, Belgium

Email for correspondence: luciano.floridi@yale.edu

**Abstract**

The emergence of Agentic Artificial Intelligence (AAI) systems capable of independently initiating digital interactions requires a new optimisation paradigm designed explicitly for seamless agent-platform interactions. This article introduces Agentic AI Optimisation (AAIO) as an essential methodology for ensuring effective integration between websites and agentic AI systems. As Search Engine Optimisation (SEO) has shaped digital content discoverability, AAIO can define interactions between autonomous AI agents and online platforms. By examining the mutual interdependence between website optimisation and agentic AI success, the article highlights the virtuous cycle that AAIO can create. It further explores the governance, ethical, legal, and social implications (GELSI) of AAIO, emphasising the need for proactive regulatory frameworks to mitigate potential negative impacts. The article concludes by affirming AAIO's essential role as part of a fundamental digital infrastructure in the era of autonomous digital agents, advocating equitable and inclusive access to its benefits.

**Acknowledgements:** The authors declare no conflicts of interest.

## 1. Introduction: From SEO to AAIO

Over the past decades, Search Engine Optimisation (SEO) has significantly influenced digital content structure, discoverability, and consumption (Enge, Spencer, and Stricchiola 2023). SEO practices have been central to enhancing website visibility, accessibility, and user engagement, effectively becoming an indispensable aspect of digital publishing and marketing strategies (Forbes Technology Council November 6 2024). The arrival of new Artificial Intelligence (AI) systems has enhanced the quality and impact of SEO and SEM (Search Engine Marketing) (Yuniarthe 2017, Rismay March 6 2023). However, with the rapid advancement of AI toward autonomous, self-directed operation—commonly termed "agentic AI" (AAI) (Acharya, Kuppan, and Divya 2025)—a parallel need arises for a new optimisation paradigm explicitly designed for AI agents. This emerging approach, identified here as Agentic AI Optimisation (AAIO), extends traditional SEO principles—such as structured data usage, metadata tagging, and content accessibility—to accommodate the unique operational characteristics of autonomous AI agents, including contextual "understanding", proactive interaction, and dynamic decision-making. Unlike SEO, which primarily enhances discoverability for human users through search engines, AAIO explicitly optimises content for autonomous artificial agents (Floridi 2025), addressing both human and machine interpretability (Baeza-Yates and Ribeiro-Neto 2011, Floridi 2019). The concept of AAIO is novel in several respects. First, while previous work has addressed either the technical design of agent-accessible digital environments (Deng et al. 2023, Zhou et al. 2024) or the governance of AI agents themselves (Kolt 2025), AAIO uniquely positions digital optimisation as a necessary infrastructural layer that mediates between agentic AI capabilities and the digital content upon which they operate. Second, AAIO differs from conventional accessibility standards and web optimisation paradigms by emphasising bidirectional, proactive interaction patterns rather than passive content retrieval. Where SEO and related frameworks assume a human end-user initiating queries, AAIO anticipates agents that autonomously navigate, synthesise, and act upon digital content with minimal human intervention. Third, AAIO addresses a critical gap identified in recent agent research: the observation that agent performance is constrained not only by model capabilities but also by the structure and semantic richness of the environments in which they operate (Kapoor et al. 2024). By articulating AAIO as a distinct optimisation paradigm, this article provides developers with actionable design principles for agent-ready digital infrastructure, offers scholars a conceptual framework for analysing the co-evolution of digital environments and autonomous systems, and supplies policymakers with the right level of abstraction through which to anticipate regulatory challenges arising from agent-mediated digital interactions.

This article argues that AAIO is not merely advantageous but essential for facilitating the successful integration of agentic AI within digital ecosystems. By structuring websites and digital content for straightforward "interpretation" and use by autonomous agents, AAIO enhances agentic AI performance, creating a mutually reinforcing relationship between digital optimisation and AI effectiveness.

The rest of the article is organised as follows. Section 2 provides a brief, conceptual overview of agentic AI, establishing its relevance and potential impact, and critically assessing barriers to its widespread adoption. Section 3 investigates the reciprocal relationship between AAIO and agentic AI, outlining some practical methodologies for optimising digital environments for autonomous agents. Section 4 highlights some pressing GELSI (governance, ethical, legal, and social implications) associated with AAIO, emphasising the necessity of proactive regulation and comprehensive legal frameworks in anticipation of future developments. In Section 5, the article concludes by emphasising the importance of AAIO as infrastructure for future digital interactions, highlighting the need to ensure equitable access to the benefits of agentic AI.

## 2. Conceptualising Agentic AI: Significance and Adoption Challenges

Agentic AI systems significantly advance AI capabilities, as they are characterised by autonomy, decision-making capacity, and adaptive responsiveness to dynamic digital environments (Ductan, January 15, 2025). Recent scholarship has sought to clarify the conceptual foundations of this emerging paradigm. Sapkota et al. (2025) distinguish between AI Agents—modular systems driven by large language models for task-specific automation—and Agentic AI systems, which represent a paradigm shift marked by multi-agent collaboration, dynamic task decomposition, persistent memory, and coordinated autonomy. This distinction is significant for AAIO, as the optimisation strategies required may differ substantially depending on whether the interacting system is a single-purpose agent or a coordinated multi-agent ensemble. Unlike conventional AI applications, which typically operate within narrow parameters and require explicit human direction, agentic AI can engage proactively with digital platforms, autonomously performing tasks and making contextually informed decisions (Edwards February 4 2025). Recent empirical work has further emphasised that agent development is driven by benchmarks, yet current evaluation practices often focus narrowly on accuracy without adequate attention to cost, reproducibility, and real-world applicability (Kapoor et al. 2024). This observation underscores the importance of AAIO: optimising digital environments for agents is not merely a matter of enabling agent functionality but of supporting the development of agents that are genuinely useful in practical contexts. A simple example may help to clarify the point.

Imagine a mobile app that schedules your daily tasks, meetings, and personal errands. Instead of following a static calendar or a fixed set of to-dos, it continuously learns and adapts based on your changing availability and priorities. To do that, the app monitors your behaviour—e.g., recording how long daily tasks usually take—it has high-level objectives—e.g., maintaining a balanced workload per day—it reacts to unexpected events, and makes trade-off decisions on your behalf—e.g., postponing less critical tasks when an urgent setback arises. The dynamic way of operating enables it to shuffle tasks around, slot in breaks when you need to recharge, or suggest better times in your schedule to manage more demanding tasks. An app of this kind would embody AAI through a key feature often described as "open-endedness" (Acharya, Kuppan, and Divya 2025), meaning there is no single prescribed way to accomplish each task. Instead of needing step-by-step instructions, the app autonomously makes day-to-day decisions on your behalf, possibly alerting you when it implements major schedule changes.

Notably, AI agents can operate in a multi-agent environment, as seen in the case of AutoGPT, an open-source AI platform built on OpenAI's GPT-4 that automates multi-step projects and workflows (Belcic, October 15, 2024). It proactively interacts with digital platforms—calling Application Programming Interfaces (APIs), browsing the web, and managing files—to complete tasks typically performed by humans (Yang,

Yue, and He 2023). For instance, when planning a trip, AutoGPT can search for flights and hotels, compare prices, and even attempt bookings without human intervention. The ability to interface with external systems and execute complex workflows makes it a truly agentic AI that acts on the user's behalf in the digital world.

The significance of agentic AI lies in its potential to transform digital interactions fundamentally, automating complex, repetitive tasks and thereby enhancing human productivity considerably. Consider that in February 2025, there were approximately 8.4 billion voice-based virtual assistants (this includes Alexa and Siri, for example), which is more than the world's population (8.2 billion), with about 1 out of 5 people worldwide using voice search (Kumar, February 4, 2025). These assistants are a likely to become AI-enabled agents. The agentic version of Alexa, called Alexa+, was announced in February 2025 (Wiggers, February 26, 2025).

In theory, AAI promises to increase efficiency in information retrieval, product selection, personalisation, and user experience optimisation, potentially reshaping entire industries (Czech September 9 2024). In practice, the adoption of AAI faces considerable challenges. Technological hurdles, such as the current limitations of Natural Language Processing (NLP) and the transparency of decision-making, raise questions regarding the reliability and trustworthiness of such systems (Belovich, February 12, 2025). There are also concerns around fully autonomous AI agents, which can create and execute new code; this raises significant safety concerns, among others (Mitchell et al. 2025). Moreover, user acceptance remains uncertain due to some well-known problems regarding privacy, transparency, and accountability. As in the past, the solutions will mirror established principles from traditional AI governance, including transparent algorithmic auditing, explicit consent mechanisms, and clear accountability frameworks, which will be essential for successfully integrating agentic AI systems.

Regulatory ambiguity further complicates adoption, as existing frameworks inadequately address the novel intricacies of autonomous digital interactions (Sassoon March 6 2025). Kolt (2025) provides a systematic analysis of these governance challenges, relying on agency law and economic theory to identify problems arising from AI agents, including information asymmetry, discretionary authority, and loyalty conflicts. Critically, Kolt argues that conventional mechanisms for addressing agency

problems—incentive design, monitoring, and enforcement—may prove ineffective for governing AI agents that make uninterpretable decisions and operate at unprecedented speed and scale. This analysis has direct implications for AAIO: optimisation practices that enhance agent capabilities without corresponding governance infrastructure may exacerbate rather than mitigate these challenges. Thus, while the potential of agentic AI is substantial, careful consideration of these adoption barriers is vital. Successfully navigating these challenges will significantly influence the ultimate realisation of agentic AI's transformative benefits. One of the initial challenges is represented by the environment in which agentic AI operates. Elsewhere (Floridi 2023b), it is argued that the success of AI depends not on its intelligence, which is non-existent, but on how well the environment is structured (*enveloped*, for engineers, *re-ontologized*, for philosophers) around its mindless agency (Floridi 2023a). Is the infosphere optimised to make Agentic AI successful? This is the question addressed in the next section.

### 3. The Mutual Reinforcement of AAIO and Agentic AI: a Virtuous Cycle

The effectiveness and success of agentic AI systems depend on the quality and structure of the digital content they interact with (Floridi and Illari 2014). This principle has been empirically validated by recent computer science research on web agents. Deng et al. (2023) introduced Mind2Web, the first dataset for developing and evaluating generalist agents for the web, showing that agent performance is fundamentally constrained by the complexity and structure of real-world websites. Their work revealed that raw HTML from real-world webpages often contains thousands of elements that are infeasible for language models to process directly, necessitating preprocessing and filtering strategies that AAIO practices could obviate through better-structured content. Similarly, Zhou et al. (2024) developed WebArena, a realistic and reproducible web environment designed to facilitate the development of autonomous agents, highlighting the gap between simulated environments and the complexities that agents encounter in authentic digital settings. These benchmarks underscore that agent success depends not only on model capabilities but also on environmental affordances—precisely the domain AAIO seeks to address. In turn, as agentic AI systems become increasingly prevalent, digital content providers have a growing incentive to optimise their offerings specifically for autonomous agents, creating a virtuous cycle of mutual enhancement (Ductan January 15 2025). But how should this optimisation happen? We observed that digital content providers, particularly websites, have enhanced visibility and improved ranking in search results through SEO. By targeting relevance, authority, and user experience—using keyword optimisation, meta tags, structured data, mobile-friendliness, and crawlability—SEO helps attract organic (non-paid) traffic (Yalçın and Köse 2010). Yet, one of the primary abilities of agentic AI is to act and interact autonomously in a digital environment, going beyond straightforward search results to seek the most structured, semantically rich data. In this way, AAI systems respond to user queries. They may employ advanced parsing techniques, such as NLP, and look for context-rich insights or real-time data, like dynamic pricing. While traditional SEO techniques are still valuable, they often focus on signals designed for conventional search engines (e.g., keyword density and meta tags), which may not suffice for agentic AI that relies on different (and deeper) forms of information quality. Consequently, web content must be optimised with an expanded approach that addresses richer metadata, robust APIs, and real-time updates to ensure accurate machine interpretation and decision-making.

AAIO incorporates several targeted strategies explicitly designed to enable efficient, accurate, and meaningful interactions by agentic AI systems. These methodologies include structured data schemas (e.g., JSON-LD,[1] RDFa[2]) to provide explicit semantic context; optimisation for NLP and voice search technologies; ensuring rapid content load times and mobile-first responsive design; and developing robust, standardised APIs for seamless integration with autonomous agents. These strategies collectively ensure that digital environments are not only easily navigable by autonomous agents but also contextually rich and reliably interpretable.

Structured data and schema markup provide explicit, machine-readable context, enabling autonomous agents to interpret and categorise digital content accurately (Pakanati March 17 2025). Natural language and voice search optimisation further facilitate intuitive interactions between agents and digital platforms, mirroring human communication patterns and ensuring greater efficacy in information retrieval and task execution (Belovich February 12 2025).

---

[1] JSON-LD (JavaScript Object Notation for Linked Data) is a way to encode linked data using the familiar JSON format, making it easier for machines to understand and process structured data on the web, particularly for search engines.

[2] RDFa (Resource Description Framework in Attributes) is a W3C recommendation that allows embedding rich metadata within HTML, XHTML, and XML documents using attributes, enabling machines to understand the meaning of web content.

Technical optimisation—including clear, logical website architecture, mobile responsiveness, rapid load times, and comprehensive metadata—also directly enhances the ability of agentic AI systems to navigate, index, and utilise digital content efficiently (Belovich February 12 2025). Similarly, robust, well-documented APIs and accessible interfaces ensure the seamless integration of autonomous agents with diverse digital platforms, thereby reinforcing their reliability and effectiveness (Forbes Technology Council, November 6, 2024). The importance of structured interfaces for agent performance has been shown by recent research on tool and API use. Yao et al. (2023) introduced ReAct, a paradigm that enables language models to generate both reasoning traces and task-specific actions in an interleaved manner, demonstrating that agents' ability to interface with external sources such as APIs and knowledge bases is critical for overcoming hallucination and improving interpretability. Patil et al. (2024) developed Gorilla, a model specifically fine-tuned for API calls that surpasses GPT-4 on this task, showing that the quality and structure of API documentation directly shapes agent behaviour and substantially mitigates hallucination. Their work shows that when agents are trained with retriever-aware methods on well-documented APIs, they exhibit strong capability to adapt to documentation changes and generate accurate calls. These findings provide empirical support for AAIO's emphasis on API-level optimisation: structured, well-documented, and semantically rich interfaces are not merely convenient but essential for reliable agent operation.

Beyond bespoke APIs, the Model Context Protocol (MCP) standardises how models discover and invoke external tools, resources, and prompts via a host–client–server design, shifting tool use from static integration to dynamic orchestration and directly enlarging the action space agents can select from (Anthropic, 2024; Hou et al., 2025; Singh et al., 2025). Its diffusion, with over 8 M weekly SDK downloads and integrations across major platforms, means AAIO must assume MCP-shaped traffic and strengthen the systems that describe and catalogue tools, since these records now shape how models act (Hasan et al., 2025; Radosevich & Halloran, 2025).

As in the SEO case, technical optimisation should also balance the desires of digital platform operators to exclude agentic AI systems. For example, SEO offers technical measures such as robots.txt, a well-known file that states which user-agents[3] can access or crawl a site.[4] However, AI companies frequently overlook these files when scraping training data and retrieving search results for search agents (Paul, June 21, 2024; Mehrotra and Marchman, June 19, 2024). This practice can be modernised for enforcement so that websites can make granular specifications about which parts of a site an agentic AI system may access.

A significant development in AAIO implementation is the "LLMs.txt" standard, which addresses a critical limitation faced by large language models: context windows that are too small for entire websites and difficulties in converting complex HTML into LLM-friendly plain text (Howard, September 3, 2024). The specification enables website owners to provide structured, agent-accessible content through a standardised markdown file, and has already been adopted by major AI companies including Anthropic and Perplexity (Garner, March 28, 2025). This is evidence of the virtuous cycle described earlier: increasing AI system adoption encourages more website implementation, enhancing both accessibility and agent effectiveness. Orchestration is also moving agent-to-agent: the Agent2Agent (A2A) protocol enables secure, auditable cooperation between agents, while OpenAI's Agents SDK incorporates MCP directly, linking discovery, action selection, and resource access in a single stack (Surapaneni et al., 2025; OpenAI, 2025; The Verge, 2025). Optimising for agent-to-agent (A2A) interactions expands AAIO beyond how agents connect to platforms, toward how they connect with each other, requiring methods to track origins, maintain shared records, and log events across their exchanges (Surapaneni et al., 2025; OpenAI, 2025).

Finally, content optimisation through regular updates, precise targeting of user intent, and leveraging AI-driven analytics for continuous content improvement ensures sustained relevance for both human users and autonomous agents (Edwards

---

[3] User-agents are a charactistic string sent by a machine in a web request that lets the receiving server know the requestors application, operating system, vendor, and/or version. AI agents making web requests like OpenAI's ChatGPT will have a unique user-agent string compared to a user with their Chrome web browser (Mozilla Developer Network, 2025).

[4] https://www.cloudflare.com/learning/bots/what-is-robots-txt/

February 4 2025). Collectively, these practices form the core of AAIO, creating an environment conducive to optimised agentic AI performance. Figure 1 compares the main aspects of SEO and AAIO.

| Aspect | Search Engine Optimisation (SEO) | Agentic AI Optimisation (AAIO) |
|---|---|---|
| *Primary Objective* | Enhance the visibility and discoverability of websites for human users via search engines. | Facilitate efficient interaction and usability of websites for autonomous AI agents. |
| *Target Audience* | Human users interacting via search engines. | Autonomous AI agents capable of independent interactions and decisions. |
| *Optimisation Techniques* | Keyword optimisation, backlinking, human-readable metadata, structured data (limited to search visibility). | Advanced structured data (JSON-LD, RDFa), NLP optimisation, API integration, and voice search compatibility. |
| *Interaction Pattern* | Passive: users initiate queries; websites respond passively. | Proactive: autonomous agents initiate tasks, actively navigate, and utilise content dynamically. |
| *Content Interpretation* | Primarily focused on keyword relevance and human search intent. | Emphasises contextual understanding, semantic clarity, and proactive intent prediction. |
| *Technological Dependencies* | Search engine indexing algorithms (e.g., Google, Bing), restriction or allow the listing of specific indexing agents. | Autonomous agent capabilities in NLP, semantic reasoning, decision-making algorithms, restriction, or allow the listing of specific AI systems. |
| *Ethical and Regulatory Issues* | Primarily, these include privacy, intellectual property, and transparency in content indexing. | Expanded concerns include algorithmic accountability, transparency in autonomous decision-making processes, mitigation of biases, safety, reliability, and compliance with data privacy regulations. |
| *Potential Societal Impacts* | Digital visibility inequality, content quality and reliability. | Broader impacts include employment displacement, digital equity, surveillance, and algorithmic bias. |

Figure 1 Comparative analysis of key aspects distinguishing SEO from AAIO.

## 4. Addressing the Governance, Ethical, Legal, and Social Implications (GELSI) of AAIO

AAIO may introduce significant governance, ethical, legal, and social implications, necessitating proactive anticipation and management.

Governance issues include establishing standards, ensuring cross-platform interoperability, implementing effective oversight mechanisms, and maintaining transparency in autonomous digital interactions (Sassoon March 6 2025). Alignment with international AI standards (e.g., ISO/IEC JTC 1/SC 42 on Artificial Intelligence) and the creation of new ones could help establish interoperable and universally accepted frameworks for AAIO practices. By way of illustration, traditional SEO is vulnerable to "SEO spam", where content is optimised not for quality, but to rank higher in search results, degradating text quality across search engines (Bevendorff et al., 2024). Similarly, AAIO could be vulnerable to equivalent tactics that "lure" agents to sites for traffic benefits. Agentic AI could also automate SEO spam and address the existing problem of AI content mills, which are designed to gain clicks with low-quality content (Knibbs, 2025).

The potential automation of SEO spam is emblematic of a wider concern related to the virtuous cycle mentioned in Section 3, where digital content providers optimise their offerings specifically for autonomous agents. Although the outcomes are not always positive, in the current SEO context, humans remain, to an extent, in the loop. Humans can decide to pay for their company to be featured at the top of sponsored results, employ content strategists to align with search algorithms, or implement technical optimisations based on documented best practices. AAIO threatens to remove humans entirely from this process, creating a scenario where AI systems generate content optimised for other AI systems with little or no human intervention. This double-exclusion is ethically risky because agentic AI functions as a hybrid of search engine and recommender system—not merely locating information but actively curating, synthesising, and presenting it with apparent authority and conviction. Users are unlikely to be given a range of results for their own evaluation, but a single, authoritatively-looking result tailored to what the agent infers the users' needs may be. If these influential recommendations are themselves products of non-transparent AAIO practices, users face unprecedented vulnerability to manipulation without their knowledge.

AAIO raises these concerns more acutely than prior optimisation paradigms for several reasons. First, unlike traditional SEO where content is ultimately presented to human users who can exercise critical judgment, AAIO-optimised content may be processed entirely by agents that lack the capacity for such judgment, amplifying the consequences of misleading or manipulative optimisation. Second, the opacity of agent decision-making processes—a challenge extensively documented in recent governance scholarship (Kolt 2025)—compounds the difficulty of detecting when AAIO practices cross from legitimate optimisation into manipulation. Third, the speed and scale at which agents operate means that problematic AAIO practices could proliferate before human oversight mechanisms can respond effectively.

However, several countervailing forces may help mitigate these risks without necessitating heavy-handed regulation. Market incentives may favour transparency, as users are likely to prefer agents that can explain their recommendations and disclose when content has been optimised for agent consumption. Industry standard-setting bodies—building on existing frameworks like robots.txt and emerging protocols like LLMs.txt and MCP—may develop norms that distinguish legitimate AAIO from manipulative practices. Additionally, contestability mechanisms, whereby users can challenge or override agent recommendations, could preserve human agency even within heavily agent-mediated environments. The challenge for governance frameworks will be to support these organic corrective mechanisms while establishing baseline protections against the most serious harms.

This creates a considerable tension between the need to optimise for users and the protection of their autonomy, and the need to optimise for agents to ensure their utility, usability, and effectiveness. Therefore, ethical considerations, especially transparency, accountability, user trust and the potential for biases in decision-making processes, are paramount in agentic AI contexts. Ensuring agents' decisions are explainable and auditable becomes a moral and societal imperative, safeguarding users' rights and mitigating algorithmic bias (Edwards February 4 2025). Furthermore, ethical frameworks must explicitly address the potential misuse of agentic AI for misinformation, manipulation, or surveillance, proactively establishing clear normative guidelines and accountability measures to protect individuals and society from adverse impacts.

From a legal perspective, AAIO raises several issues, particularly regarding Intellectual Property (IP) and data use regulations. When AI agents access, read, and act upon website information, they may extract content in ways that implicate legal rights—i.e., IP-generated content. Although content providers—e.g., website owners—may deliberately optimise this information for agents through structured data feeds or APIs, they should clearly establish usage rights through explicit terms of service, as automated scraping is often restricted and usually requires obtaining an API license explicitly agreed upon by both parties. Content owners commonly implement tiered protection strategies—e.g., a news site might allow AI agents to access headlines or summaries while restricting access to complete articles without appropriate licensing arrangements. The regulatory landscape governing these interactions varies by jurisdiction. The European Union's (EU) framework on text-and-data mining (Directive (EU) 2019/790 on Copyright in the Digital Single Market), for instance, allows AI systems to scrape content by default unless the content owner opts out. This places the burden on European website operators to actively opt out if they wish to restrict AI agents' access. These considerations imply that effective AAIO implementation may necessitate the development of a formalised API or dedicated data feed for AI agents. Such contractual arrangements would enable website owners to set the terms (permitted uses, rate limits, etc.). Conversely, major uncertainty arises, leaving the ultimate word to the courts. Once the IP terms are defined, AI developers and content owners should communicate—e.g., use standard protocols for bot access—to ensure those terms are complied with.

Other significant legal issues concern the user's privacy and data protection, given that AI agents interact with digital platforms on the user's behalf. For example, when optimising for an AI agent that books flights, a website would need to implement structured authentication endpoints that verify the agent's authorisation to access sensitive payment information, such as credit card details and credit line data from banking institutions. The banking website, in turn, requires AAIO-specific verification protocols to confirm that the AI agent is authorised to initiate the transaction and proceed with payment. Under the General Data Protection Regulation (GDPR), websites that collect data through agent-optimised interfaces become data controllers with specific compliance obligations regarding the structured data they make available to AI systems. This requires implementing specialised metadata schemas that clearly delineate which personal data fields agents can access and under what conditions. If a website collects personal data during AI agent interactions—such as bank account numbers during a transaction—it assumes data controller responsibilities and must optimise its interfaces to provide appropriate safeguards. Website optimisation for ADMT (Automated Decision-Making Technology) compliance under the California Consumer Privacy Act (CCPA) similarly necessitates implementing machine-readable consent indicators and transparent operation logs documenting agent activities.

Effective AAIO requires designing privacy-by-design elements specifically for agent interactions, including granular permission structures, real-time consent verification endpoints, and machine-readable data minimisation protocols. These specialised optimisation techniques not only support regulatory compliance but also enhance user trust when agents interact with optimised websites. Without such optimisations, websites risk creating compliance vulnerabilities if their agent-focused interfaces inadvertently bypass user opt-out mechanisms or fail to implement appropriate user oversight for different types of agent-initiated transactions.

These are just some legal issues that AAIO may trigger. It is early days, but one can already expect that robust and comprehensive regulatory and governance frameworks will have to be developed to clarify complex issues around data privacy (e.g., compliance with GDPR and CCPA), intellectual property ownership and usage rights regarding AI-generated content, liability assignment for autonomous decision-making processes, and algorithmic accountability.–In general, current legal statutes appear to be insufficiently equipped to govern the interactions of autonomous agents. Proactive regulatory measures must clearly define standards for agentic AI liability, transparency requirements, data privacy compliance, and mechanisms for accountability and redress (Belovich February 12 2025).

The social implications of AAIO encompass a broader range of potential issues, including employment displacement, changes in user-agent interaction dynamics, and digital inequality. For instance, the widespread adoption of AAIO may intensify trust and dependency relationships with AI systems, while the progressive atrophy of critical evaluation skills occurs as users increasingly delegate complex decision-making processes to algorithmic agents. Concurrently, a concerning socioeconomic dimension emerges as differential access to sophisticated agentic AI technologies threatens to create unprecedented stratification patterns in digital societies. It is a new version of the digital divide, as the technological asymmetry could transform differential access to advanced AI capabilities into a significant determinant of socioeconomic opportunities within increasingly AI-mediated environments, potentially exacerbating existing social inequalities.

Addressing these challenges requires coordinated intervention strategies from both policymakers and industry stakeholders, including targeted educational initiatives, comprehensive retraining programs, public awareness campaigns, and deliberate access equalization mechanisms to ensure that AAIO's benefits are distributed according to principles of fairness and inclusivity across diverse societal segments (Forbes Technology Council November 6 2024).

## 5. Conclusion: AAIO as Essential Infrastructure

As digital interactions increasingly involve autonomous agents, optimising websites and online resources tailored to their needs becomes an essential infrastructure component. AAIO not only enables the effective deployment of agentic AI but also enhances overall digital functionality and usability. However, the successful implementation of AAIO requires careful and proactive management of governance, ethical, legal, and social implications. Comprehensive regulatory frameworks, technical standards, transparency initiatives, and public education programmes must be designed and implemented collaboratively to ensure equitable access and mitigate negative consequences.

Ultimately, AAIO represents a significant evolution in digital optimisation practices, essential for fully realising agentic AI's potential. For agentic AI to succeed, its environment must be deliberately optimised and supportive, analogous to designing the right ecosystem for a new organism.

Ensuring equitable and inclusive access to AAIO and the subsequent widespread benefits of agentic AI will be crucial for mitigating digital divides and preventing the exacerbation of existing societal inequalities. Policymakers, industry leaders, and civil society must collaborate to develop education programmes, skill retraining schemes, and accessible infrastructure, ensuring that diverse populations can benefit equally from advancements in agentic AI technologies. Fostering transparency, accountability, and inclusivity in AAIO practices will be essential for maintaining societal trust, democratic accountability, and broader acceptance of this transformative technological shift (Eubanks 2018, van Dijk 2020). Of course, none of this may happen, prevention being something we tend to neglect rather than implement, but it is worth trying.

**Acknowledgements**

The authors declare that they have no conflict of interest.